\documentclass[sn-mathphys-num]{sn-jnl}

\usepackage{graphicx}%
\usepackage{multirow}%
\usepackage{amsmath,amssymb,amsfonts}%
\usepackage{amsthm}%
\usepackage{mathrsfs}%
\usepackage[title]{appendix}%
\usepackage{xcolor}%
\usepackage{textcomp}%
\usepackage{manyfoot}%
\usepackage{booktabs}%
\usepackage{algorithm}%
\usepackage{algorithmicx}%
\usepackage{algpseudocode}%
\usepackage{listings}%
\usepackage{rotating}

\usepackage{adjustbox}
\usepackage[inline]{enumitem}
\usepackage{tablefootnote}
\newcommand{\mathbbm}[1]{\text{\usefont{U}{bbm}{m}{n}#1}}
\newcommand{\fmacro}{F1{\textsubscript{macro}}}

\theoremstyle{thmstyleone}%
\theoremstyle{thmstyletwo}%

\theoremstyle{thmstylethree}%

\begin{document}

\title[Article Title]{Weakly Supervised Veracity Classification with LLM-Predicted
Credibility Signals}


\author{\fnm{João A.} \sur{Leite}}\email{jaleite1@sheffield.ac.uk}

\author{\fnm{Olesya} \sur{Razuvayevskaya}}\email{o.razuvayevskaya@sheffield.ac.uk}

\author{\fnm{Kalina} \sur{Bontcheva}}\email{k.bontcheva@sheffield.ac.uk}

\author{\fnm{Carolina} \sur{Scarton}}\email{c.scarton@sheffield.ac.uk}

\affil{\orgdiv{Department of Computer Science}, \orgname{The University of Sheffield} \\ \orgaddress{\street{Regent Court, 211 Portobello Street}, \city{Sheffield}, \postcode{S1 4DP}, \country{United Kingdom}}}


\abstract{Credibility signals represent a wide range of heuristics typically used by journalists and fact-checkers to assess the veracity of online content. Automating the extraction of credibility signals presents significant challenges due to the necessity of training high-accuracy, signal-specific extractors, coupled with the lack of sufficiently large annotated datasets. This paper introduces \textsc{Pastel} (\textbf{P}rompted we\textbf{A}k \textbf{S}upervision wi\textbf{T}h cr\textbf{E}dibility signa\textbf{L}s), a weakly supervised approach that leverages large language models (LLMs) to extract credibility signals from web content, and subsequently combines them to predict the veracity of content without relying on human supervision. We validate our approach using four article-level misinformation detection datasets, demonstrating that \textsc{Pastel} outperforms zero-shot veracity detection by $38.3\%$ and achieves $86.7\%$ of the performance of the state-of-the-art system trained with human supervision. Moreover, in cross-domain settings where training and testing datasets originate from different domains, \textsc{Pastel} significantly outperforms the state-of-the-art supervised model by $63\%$. We further study the association between credibility signals and veracity, and perform an ablation study showing the impact of each signal on model performance. Our findings reveal that $12$ out of the $19$ proposed signals exhibit strong associations with veracity across all datasets, while some signals show domain-specific strengths.}




\maketitle

\section{Introduction}\label{sec:introduction}

In the era of rapidly spreading mis- and disinformation\footnote{Since veracity classification is concerned with determining whether online content is true or false, this paper will use the terms misinformation and disinformation interchangeably, as they both involve false content, and veracity classification does not consider the intent of the user sharing the false content.}, the task of automatic veracity classification of online content has emerged as a prominent field of research \cite{zhoufakesurvey2020}. Despite significant progress, several limitations and challenges persist. State-of-the-art methods typically rely on supervised learning, and thus require high-quality, manually annotated datasets. The creation of such datasets is time-consuming, and the evolving nature of misinformation necessitates the continuous development of new datasets \citep{fu2023feature, 9207498, kdmile}. Additionally, supervised methods often struggle to generalise across different misinformation domains (e.g., politics and celebrity gossip), resulting in considerable decrease in performance if in-domain data is unavailable \cite{perez-rosas-etal-2018-automatic, 9418411}. 

To address these issues, prior work has employed weakly supervised methods that leverage indirect learning signals to classify the veracity of content without relying on annotated data. Current methods use weak signals as a combination of simple syntactic features (e.g., count of words) and user engagement with the misinformation content (e.g., number of shares) \cite{shu2020weaksocialsupervision, helmstetter2018weakly, Wang_Yang_Ma_Xu_Zhong_Deng_Gao_2020}. The latter is particularly ineffective, as models that depend on engagement features require the content to be spread and interacted with before the model can accurately detect its deceptive nature, by which time the misinformation narrative has already caused harm. In spite of the simplicity of the aforementioned signals, the challenge of integrating more sophisticated signals (e.g. credibility signals\footnote{See the report by \citet{w3c-signals} for an overview of credibility signals.} defined by experts) poses a paradox: complex signals demand specialised models and annotated datasets for accurate extraction \cite{dimou2022evaluating}, which undermines the premise of employing weak supervision in the first place.

Pretrained large language models (LLMs) offer promising opportunities to address the aforementioned challenges. While further research is necessary to fully understand their potential and limitations, LLMs have demonstrated remarkable zero-shot performance in various NLP tasks, including common sense reasoning, reading comprehension, and closed-book question answering 
\cite{touvron2023LLaMa}, at times even surpassing state-of-the-art supervised approaches \cite{NEURIPS2020_1457c0d6}. LLMs exhibit strong recall of factual knowledge without fine-tuning \cite{petroni-etal-2019-language}, suggesting that the external knowledge acquired during pretraining could be harnessed to extract complex signals from textual content without requiring further fine-tuning with annotated datasets.

Our contribution with this work is the proposal of Prompted weAk superviSion wiTh crEdibility signaLs (\textsc{Pastel}), \textbf{an approach modelled on the verification process typically adopted by journalists and fact-checkers}, who assess the veracity of online content \textbf{using a wide range of credibility signals}. We leverage the task-agnostic capabilities of LLMs to extract nineteen sophisticated credibility signals from news articles in a zero-shot setting (i.e., without training the model with ground truth labels). These signals are then aggregated into a binary (\textit{misinformation/non-misinformation}) veracity label using weak supervision. 

Our comprehensive experiments demonstrate that \textsc{Pastel} \textbf{outperforms zero-shot veracity classification by $38.3\%$, and attains $86.7\%$ of the performance of the state-of-the-art supervised model}, which relies on domain-specific training data. Moreover, \textsc{Pastel} \textbf{outperforms the state-of-the-art supervised model by $63\%$ in cross-domain settings}, underscoring its applicability to real-world scenarios where misinformation rapidly evolves and domain-specific training data is limited. Lastly, we investigate the role of each credibility signal in predicting content veracity by inspecting their statistical association with the human-annotated veracity labels, and through an ablation study in which \textsc{Pastel}'s performance is measured after individual signals are removed. Our analysis provides valuable insights highlighting the importance of domain-specific credibility signals, and how a diverse range of credibility signals is key in enhancing the model’s performance.

The remainder of this paper is structured as follows: Section~\ref{sec:related_work} presents an overview of relevant previous work. Section~\ref{sec:pwscs} describes our proposed method. The experimental setup is presented in Section~\ref{sec:experimental_setup}, whilst results are discussed in Section~\ref{sec:results}. In Section~\ref{sec:signals_analyses}, we analyse and discuss the predicted credibility signals. Section~\ref{sec:discussion_conclusion} presents a discussion of implications of this work, points to future work, and makes concluding remarks. We make our code and data fully available \cite{pastel-repo}.

\section{Related Work} \label{sec:related_work}

\subsection{Article-level Veracity Classification} \label{sec:related_work_datasets}
Building models to automatically assess content veracity generally relies on human-annotated datasets. Most benchmark corpora focus on short claims \citep{vlachos2014fact, ferreira2016emergent, wang-2017-liar, thorne-etal-2018-fever} or social media data such as Facebook posts \citep{potthast-etal-2018-stylometric, santia2018buzzface, tacchini2017some}, tweets \citep{zubiaga2016analysing, mitra2015credbank, 10.1145/3219819.3219903}, and Reddit threads \cite{nakamura-etal-2020-fakeddit}. However, article-level veracity assessment relies on more context and nuance, making annotation more challenging and less scalable, therefore fewer datasets are available \citep{shu2020fakenewsnet, perez-rosas-etal-2018-automatic, li2020toward, hossain-etal-2020-banfakenews}. This section describes four article-level datasets commonly employed in works studying automatic veracity detection and cross-domain generalisation. We also present the key classification approaches used.

\citet{perez-rosas-etal-2018-automatic} introduced two datasets: FakeNewsAMT and Celebrity, annotated with binary veracity labels. FakeNewsAMT contained political news articles from six topics with deceptive versions created by crowdsourced workers. The Celebrity dataset included web articles about celebrities verified against gossip-checking sites. Both datasets achieved annotation agreement scores of $70\%$ and $73\%$, respectively. Also, the authors performed a cross-domain analysis by training their best model with one of the datasets and testing on the other. Results showed a drop in performance of $13.5\%$ for the FakeNewsAMT dataset, and $34.2\%$ for the Celebrity dataset. Studies on these datasets used models such as SVMs with word embeddings, grammatical features, and word-level attention with multi-layer perceptrons \cite{saikh2020deep, gautam2020sgg}. Transfer learning models such as RoBERTa, GPT-2, XLNet, DeBERTa, and BERT surpassed feature-based methods, with the best reported \fmacro\ scores of 0.99 for FakeNewsAMT and 0.82 for Celebrity using RoBERTa. However, they struggled in cross-domain settings, dropping $40\%$ in performance \cite{9418411}. 

\citet{shu2020fakenewsnet} presented PolitiFact and GossipCop, two binary article-level datasets. PolitiFact included politically themed articles assessed by journalists, while GossipCop focused on celebrity stories verified by a rating system. Previous methods evaluated on these datasets include CNNs, knowledge-aware attention networks, and convolutional Tselin Machines \cite{shu2020fakenewsnet, dun2021kan, bhattarai-etal-2022-convtexttm}. The current state-of-the-art results were achieved by \citet{rai2022fake}, who fine-tuned BERT model, achieving \fmacro\ scores of 0.88 for PolitiFact and 0.89 for GossipCop. They experimented with an LSTM layer on top of BERT, which slightly improved performance by $0.02$ for PolitiFact but did not affect GossipCop.

\subsection{Credibility Signals}\label{sec:cred-signals-related-wk}
The term {\em credibility signals} refers to a wide range of measurable heuristics that collectively help journalists assess the overall trustworthiness of information. Examples of credibility signals include the analysis of article titles \cite{horne2017just}, writing style \cite{afroz2012detecting}, rhetorical structure \cite{rashkin2017truth, nikolaidis-etal-2024-exploring}, linguistic features \cite{o2018language}, emotional language \cite{10.1145/3331184.3331285}, biases \cite{dufraisse-etal-2022-dont}, and logical fallacies and inferences \cite{doi:10.1177/09579265221076609}. Additionally, credibility signals comprise meta-information that extends beyond the textual content of the article, such as the author’s reputation and external references \cite{sitaula2020credibility}.

The W3C Credible Web Community Group (CWCG) \cite{w3c-signals} performed the most extensive attempt to date at cataloguing credibility signals by defining and documenting hundreds of signals. \citet{dimou2022evaluating} selected 23 credibility signals defined by the W3C CWCG and built a modular evaluation pipeline for the task of predicting the credibility of content. Their signals were a mixture of (i) simple syntatic features (e.g., word length, word count, exclamation marks), (ii) metadata (e.g., author name, URL domains), and (iii) a smaller set of complex features extracted by specialised classifiers trained for each of them (e.g., sentiment, clickbait). These signals were grouped into $10$ modules, and each module was manually assigned an importance weight that defined its contribution to the overall credibility of the web page. The authors found that morphological, syntactic, and emotional features demonstrated the highest predictive capability for determining the credibility of web content.

To the best of our knowledge, the only dataset annotated with several credibility signals was introduced by \citet{10.1145/3184558.3188731}. They employed six trained annotators to label articles with $17$ different content indicators and $11$ context indicators based on the W3C CWCG definitions. However, their dataset was a feasibility study with a small sample size of only $40$ annotated articles, which severely limits its utility for training supervised machine learning models.

\subsection{Veracity Classification with Weak Supervision}
{\em Programmatic weak supervision} (PWS) is a semi-supervised learning paradigm that encodes noisy probabilistic labels using multiple {\em labeling functions} that are correlated with the objective task \citep{fu2020fast, varma2019multi, ratner2016data, smith2022language}. Several prior works applied weak supervision techniques to detect the veracity of online content. A common theme among these works was the use of social media metadata, syntactic features, and user interactions with misinformation content as weak signals. 

\citet{shu2020weaksocialsupervision} incorporated multiple weak signals from user engagements with content. Their weak signals included (i) \textit{sentiment}, which considered the average sentiment scores inferred from users sharing a given news piece; (ii) \textit{bias}, which was modelled by inspecting how closely the user’s interests matched those of people with known public biases; and (iii) \textit{credibility}, which considered the size of the cluster containing the user. This was modelled on the hypothesis that low-credibility users were likely to coordinate and form large clusters, while high-credibility users tended to form small clusters. Their best classifier trained exclusively with weak signals was a RoBERTa model that achieved an average \fmacro\ score of $0.535$ across two datasets.

\citet{helmstetter2018weakly} applied weak supervision for misinformation detection on Twitter. They used five sets of features as weak signals: (i) a total of $53$ \textit{user-level} features, such as the frequency of tweets, ratio of retweets, number of followers, etc.; (ii) a total of $69$ \textit{tweet-level} features, such as word count and the ratio of question and exclamation marks; (iii) \textit{text-level} features comprising TF-IDF encoded vectors representing the tweet text; (iv) \textit{topic-level} features consisting of automatically derived topics using LDA; and (v) \textit{sentiment-level} features representing the ratio of positive, negative, and neutral words in the text. Their best configuration used an XGBoost classifier trained with the proposed features, achieving an \fmacro\  of $0.77$ for detecting a set of misinformation tweets labelled by themselves.

\citet{Wang_Yang_Ma_Xu_Zhong_Deng_Gao_2020} proposed \textit{WeFEND}, a reinforced weakly-supervised fake news detection framework. Their approach leveraged user feedback on known misinformation articles as weak signals. They trained a classifier using these signals and applied it to predict misinformation in articles with unknown veracity, but for which user feedback was available. They evaluated their approach on a dataset of news articles published by WeChat official accounts, along with the corresponding user feedback. Their model achieved an F1-score of $0.880$ for misinformation articles and $0.810$ for non-misinformation articles.

In conclusion, our approach differs from previous works in two key aspects. First, \textsc{Pastel} does not rely on any metadata related to user engagement with the misinformation article, but operates exclusively on the textual content of the article. This distinction is crucial because models that depend on engagement features require that the content is spread and interacted with before the model can accurately detect it, by which time the misinformation narrative has already caused harm. Additionally, \textsc{Pastel} leverages signals defined by specialists from the W3C Credible Web Community Group (CWCG), which encompass more sophisticated concepts (e.g., whether the content presents evidence) compared to user engagement statistics (e.g., number of shares) or syntactic features (e.g., word count) used in previous works. To annotate these complex signals without relying on annotated data, we employ LLMs to predict the weak signals in a zero-shot setting (i.e., without any fine-tuning with annotated data).



\section{Prompted weAk Supervision wiTh crEdibility signaLs (\textsc{PASTEL})} \label{sec:pwscs}
\begin{figure*}[h]
\begin{adjustbox}{width=\columnwidth}
    \includegraphics{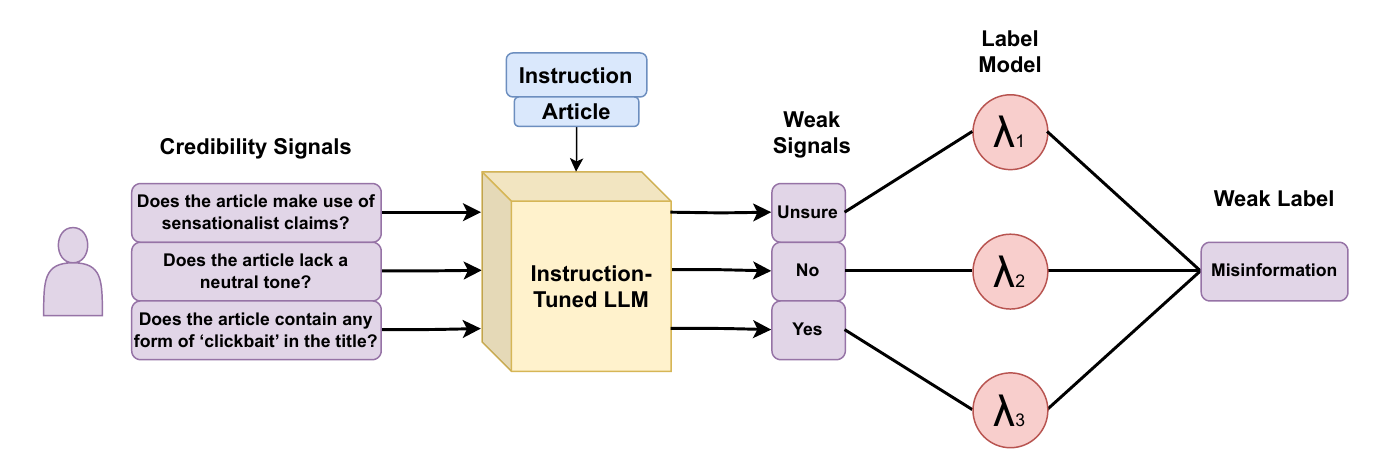}
\end{adjustbox}

  \caption{Illustration of \textsc{PASTEL}.}
  \label{fig:diagram}
\end{figure*}

\textsc{Pastel} draws inspiration from the verification practices employed by journalists and fact-checkers, who determine the truthfulness of online content based on an extensive array of credibility indicators. Our method harnesses the task-agnostic abilities of large language models (LLMs) to identify nineteen nuanced credibility signals from news articles in a zero-shot setting, meaning the model operates without training with ground truth labels. Subsequently, we integrate these signals to perform a binary classification (\textit{misinformation} or \textit{non-misinformation}) through a process of weak supervision. Figure~\ref{fig:diagram} provides an overview of the approach, illustrating it with three examples of credibility signals. In the following sections, each component is described in greater detail.

\subsection{Credibility Signals Considered}\label{sec:credibility_signals}
We leverage nineteen credibility signals, all of which have been shown to be relevant for assessing content veracity. Table~\ref{tab:full_cred_signals} displays these signals, that provide a solid foundation of well-defined and validated indicators of content credibility. Note that all our signals are formulated so that \textbf{their presence in the content indicates a lack of credibility}. 

\begin{table}[h]
\caption{Credibility signals and their respective definitions. \newline \textasteriskcentered \citet{doi:10.1177/09579265221076609} \textdaggerdbl \citet{dufraisse-etal-2022-dont} $\#$\citet{10.1145/3184558.3188731} 
\textdagger \citet{w3c-signals}}
\label{tab:full_cred_signals}
\centering
    \begin{tabular}{@{}ll@{}}
    \toprule
    \textbf{Credibility Signal} & \textbf{Definition} \\
    \midrule
    Evidence\textsuperscript{\textasteriskcentered} & \begin{tabular}[c]{@{}l@{}}Fails to present any supporting evidence or arguments to \\ substantiate its claims. \end{tabular} \\ \midrule
    Bias\textsuperscript{\textdaggerdbl} & \begin{tabular}[c]{@{}l@{}}Contains explicit or implicit biases (e.g. confirmation bias, \\ selection bias, framing bias).\end{tabular} \\ \midrule
    Inference\textsuperscript{$\#$} & Makes claims about correlation and causation. \\ \midrule
    Polarising Language\textsuperscript{$\#$} & \begin{tabular}[c]{@{}l@{}}Uses polarising terms or makes divisions into sharply contrasting \\ groups or sets of opinions or beliefs.\end{tabular} \\ \midrule
    Document Citation\textsuperscript{$\#$} & Lacks citations of studies or documents to support its claims. \\ \midrule
    Informal Tone\textsuperscript{$\#$} & Uses all caps or consecutive exclamation or question marks. \\ \midrule
    Explicitly Unverified Claims\textsuperscript{\textdagger} & \begin{tabular}[c]{@{}l@{}}Contains claims that are explicitly lack confirmation.\end{tabular} \\ \midrule
    Personal Perspective\textsuperscript{\textdagger} & Includes the author’s own personal opinions about the subject. \\ \midrule
    Emotional Valence\textsuperscript{\textdagger} & \begin{tabular}[c]{@{}l@{}}Language carries emotional valence that is predominantly negative \\ or positive rather than neutral. \end{tabular} \\ \midrule
    Call to Action\textsuperscript{\textdagger} & \begin{tabular}[c]{@{}l@{}}Contains language that can be understood as a call to action, \\ requesting readers to follow through with a particular task \\ or telling readers what to do.\end{tabular} \\ \midrule
    Expert Citation\textsuperscript{\textdagger} & Lacks citations of experts in the subject. \\ \midrule
    Clickbait\textsuperscript{\textdagger} & \begin{tabular}[c]{@{}l@{}}Title contains sensationalised or misleading headlines in order to \\ attract clicks.\end{tabular} \\ \midrule
    Incorrect Spelling\textsuperscript{\textdagger} & Contains significant misspellings and/or grammatical errors. \\ \midrule
    Misleading About Content\textsuperscript{\textdagger} & \begin{tabular}[c]{@{}l@{}}Title emphasises different information than the body topic.\end{tabular} \\ \midrule
    Incivility\textsuperscript{\textdagger} & Uses stereotypes and/or generalisations of groups of people. \\ \midrule
    Impoliteness\textsuperscript{\textdagger} & Contains insults, name-calling, or profanity. \\ \midrule
    Sensationalism\textsuperscript{\textdagger} & \begin{tabular}[c]{@{}l@{}}Presents information in a manner designed to evoke strong \\ emotional reactions. \end{tabular} \\ \midrule
    Source Credibility\textsuperscript{\textdagger} & Cites low-credibility sources. \\ \midrule
    Reported by Other Sources\textsuperscript{\textdagger} & \begin{tabular}[c]{@{}l@{}}Presents a story that was not reported by other reputable \\ media outlets.\end{tabular} \\
    \bottomrule
    \end{tabular}
\end{table}

The vast majority of the signals used in our experiments were proposed by the W3C (the Web Standards Organisation) Credible Web Community Group \cite{w3c-signals}, who defined numerous credibility indicators to help users and machines identify trustworthy content, i.e. content that is reliable, accurate, and shared in good faith (see Table \ref{tab:full_cred_signals}). The aim of our work is not to propose new credibility signals but to use those already established by subject matter experts and demonstrate how they can enhance automatic veracity classification.

\subsection{Signal Extraction (LLM Prompting)}
Instruction-tuned LLMs operate in a question-answering manner through the use of prompts. A prompt is a specific query given to the model to instruct it to perform a task. With carefully crafted prompts, the LLM’s capabilities can be harnessed to extract the credibility signals. \autoref{fig:prompt} displays the prompt template employed to extract a single credibility signal in a question-answering approach using an instruction-tuned LLM.

\begin{figure*}[!h]
\centering
  \begin{minipage}[h]{\linewidth} 
      \fbox{ 
          \parbox{\linewidth}{ 
          \#\#\# Instruction: \\
              You are a helpful and unbiased news verification assistant. You will be provided with the title and the full body of text of a news article. Then, you will answer further questions related to the given article. Ensure that your answers are grounded in reality, truthful and reliable. You are expected to answer with `Yes' or `No', but you are also allowed to answer with `Unsure' if you do not have enough information or context to provide a reliable answer. \\ \\
              \#\#\# Input: \\
              \{title\} \\
              \{text\} \\ \\
              \{question\} (Yes/Unsure/No) \\ \\
              \#\#\# Response:
          }
      }
  \caption{Prompt template to extract credibility signals.}
  \label{fig:prompt}
  \end{minipage}
  \end{figure*}

The prompt uses the Alpaca template \cite{taori2023alpaca}, and contains 3 distinct sections: `Instruction', `Input', and `Response'.
The `Instruction' section of the prompt guides the model towards extracting each credibility signal in an unbiased, grounded in reality, truthful, and reliable manner, and also ensures that the model only outputs valid answers (\textit{Yes}, \textit{No}, and \textit{Unsure}). The `Input' section is filled with the title and body of text of the input news article, followed by a question associated to a credibility signal. This is essentially a mapping from the definition of the respective signal to a question. For example, the definition for the \textit{Inference} signal (see Table~\ref{tab:full_cred_signals}) is mapped to the following question: ``Does this article make claims about correlation and causation?''. Immediately following the question, we explicitly state the three candidate answers (i.e., \textit{Yes}, \textit{Unsure}, and \textit{No}) to reinforce that the model should only output these answers. Finally, the `Response' section is left blank to allow the LLM to perform text completion. Note that this template allows for the extraction of one credibility signal at a time. Therefore, for each input news article, nineteen prompts are created, each with a different question corresponding to a distinct credibility signal. These prompts are fed sequentially to the LLM, with no additional context carried over from previous interactions.

\subsection{Weak Supervision} \label{sec:weak_supervision}
After extracting the credibility signals, our objective is to combine the signals into binary veracity labels (\textit{misinformation} or \textit{non-misinformation}). The simplest approach is to apply a majority voting heuristic: if the majority of signals are triggered, the outcome is classified as \textit{misinformation}; otherwise, it is classified as \textit{non-misinformation}. However, this approach has limitations, as all signals are treated equally, whereas ideally,
signals with higher accuracy should influence the outcome more than those with lower accuracy. Moreover, signals can be highly correlated, leading to duplicated or nearly identical outputs (i.e., double voting) which can bias the final prediction.

To address these challenges, we employ weak supervision to determine signal weights without relying on annotated data. Instead, weights are estimated from empirical statistics derived from their distribution. Our goal is to train a parameterised classification model, denoted as $h_\theta$, where, for a given news article $x \in X$, the model predicts its veracity label $y \in Y$ (where $Y \in \{0, 1\}$). In a supervised learning setting, $h_\theta$ is trained on a dataset comprising pairs of inputs and ground truth labels, denoted as $(x_{train}, y_{train})$. However, in weakly supervised learning, we lack access to $y_{train}$. Instead, we generate training labels using a set of labeling functions $\lambda: X \rightarrow Y \cup \{-1\}$, where `-1' indicates abstention (in our setting, the `Unsure' class).

Each labeling function $\lambda$ is expected to exhibit some correlation with $Y$, although they may be noisy, meaning they do not necessarily provide highly accurate predictions for $Y$ individually. Assuming we have $m$ inputs and $n$ labeling functions, $\Lambda_{ij}$ represents the output of labeling function $\lambda_j$ for input $x_i$, resulting in a matrix as follows:

\begin{equation}
    \Lambda = \begin{bmatrix}
\lambda_0(x_0) & ... & \lambda_{n-1}(x_0)\\ 
 \vdots & \ddots  & \vdots \\ 
 \lambda_0(x_{m-1}) & ... & \lambda_{n-1}(x_{m-1}) 
\end{bmatrix}_{m\times n}
\end{equation}

Next, the goal is to transform $\Lambda$ into a vector of probabilistic weak labels $\Tilde{Y} = (\Tilde{y}_0, ..., \Tilde{y}_{m-1})$, with $\Tilde{y_i} \in [0, 1]$. To do so, we train a generative model $p_\theta(\Lambda, Y)$ to obtain weights $\theta_j$ that calibrate the contribution of $\lambda_j$ towards $\Tilde{Y}$. Specifically, we use the approach by \citet{ratner2017snorkel}, which defines factor types representing the labeling propensity, and pairwise correlations between labeling functions $j$ and $k$ for the $i$th input:

\begin{equation}
\begin{split}
    &\phi^{Lab}_{i,j}(\Lambda,Y) = \mathbbm{1}\{\Lambda_{i,j} \neq -1 \} \\
    &\phi^{Corr}_{i,j,k}(\Lambda,Y) = \mathbbm{1}\{\Lambda_{i,j} = \Lambda_{i,k} \}
\end{split}
\end{equation}
The factor types are concatenated into a single vector $\phi_{i}$ for each input $x_i$, and the parameters of the model are defined as $w \in \mathbbm{R}^{2n+|C|}$, where $C$ is a set of potentially correlated pairs of labeling functions. The label model is defined by \autoref{eq:label_model}, where $Z_{w}$ is a normalising constant:
\begin{equation} \label{eq:label_model}
    p_{w}(\Lambda, Y) = Z_{w}^{-1}\exp{\left ( \sum_{i=0}^{m-1} w^{T}\phi_{i}(\Lambda, y_{i})\right )} 
\end{equation}
The model learns without access to the ground truth labels $Y$, thus the objective is to minimise the negative log marginal likelihood given the observed outputs of the labeling functions $\Lambda$:
\begin{equation} \label{eq:weak_objective}
    \hat{w}=\underset{w}{\arg \min} - \log{\sum_{Y} p_{w}(\Lambda, Y)}
\end{equation}

The trained label model is then used to infer the probabilistic weak labels $\Tilde{Y} = p_{\hat{w}}(Y|\Lambda)$, and the discrete predictions (\textit{misinformation}/\textit{non-misinformation}) are obtained by taking the $argmax$ of each weak label $\tilde{y}_i \in \Tilde{Y}$.

\section{Experimental Setup} \label{sec:experimental_setup}
In this section we describe the datasets, metrics, models, and techniques employed to assess the performance of our method in comparison to other strong baselines. The classification setting is the following: given the title and body of a news article, predict it's veracity as either \textit{misinformation} or \textit{non-misinformation}. Initially, we assess the models’ performance within the same domain, where both the train and test sets are derived from the same dataset. Subsequently, we evaluate the models’ cross-domain performance, where the train and test sets originate from different datasets.

\subsection{Datasets} \label{sec:datasets}
We experiment with four English article-level misinformation datasets: PolitiFact and GossipCop by \citet{shu2017fake}, and FakeNewsAMT and Celebrity by \citet{perez-rosas-etal-2018-automatic}. These datasets are chosen because they cover two distinct domains: GossipCop and Celebrity focus on entertainment news, whereas PolitiFact and FakeNewsAMT concentrate on politics. This distinction allows us to assess the model’s ability to generalise beyond its training data domain. Furthermore, these datasets exhibit unique characteristics that may impact model performance and are therefore crucial for evaluation.

\begin{itemize}
    \item GossipCop's classes are considerably imbalanced towards the negative class ($77.6\%$). Other datasets are near to perfectly balanced.
    \item Gossipcop has more than $10$ times the number of articles than the combination of the three other datasets.
    \item PolitiFact's average document length is considerably larger than other datasets, with $2605.2$ average tokens per article. Contrastingly, FakeNewsAMT has only $178.4$ average tokens per article, which is notably fewer than others.
\end{itemize}

Although PolitiFact and GossipCop contain additional social-context data in the form of tweets, we only use content-related attributes (title and body) as input to the models. \autoref{tab:datasets} presents the class distributions and average number of tokens for each dataset. 

\begin{table}[htb]
\caption{Datasets used throughout the experiments along with their label distributions and average number of tokens.}
\label{tab:datasets}
\centering
\begin{tabular}{@{}lccc@{}}
\toprule
\textbf{Dataset} & \multicolumn{1}{c}{\textbf{\# Misinformation}} & \multicolumn{1}{c}{\textbf{\# Non-misinformation}} & \multicolumn{1}{c}{\textbf{\# Tokens (avg.)}} \\ \midrule
PolitiFact & $308$ ($44.6\%$) & $383$ ($55.4\%$) & $2605.2$ \\
GossipCop & $3924$ ($22.4\%$) & $13596$ ($77.6\%$) & $981.3$ \\
FakeNewsAMT & $240$ ($50\%$) & $240$ ($50\%$) & $178.4$ \\
Celebrity & $250$ ($50\%$) & $250$ ($50\%$) & $635.5$ \\ \bottomrule
\end{tabular}
\end{table}

\subsection{Evaluation}
Similar to the previous works that experimented with the four datasets \cite{shu2020fakenewsnet, perez-rosas-etal-2018-automatic, dun2021kan, rai2022fake, bhattarai-etal-2022-convtexttm, saikh2020deep, gautam2020sgg, goel-2021-multidomain}, we use the \fmacro\ score as the main evaluation metric. The \fmacro\ score is defined in Equation~\ref{eq:f1macro}, where $N$ is the number of classes, and $TP_i$, $FP_i$, and $FN_i$ correspond to the number of true positives, false positives, and false negatives, respectively, for class $i$. This metric is particularly suitable for datasets with skewed class distributions, as it returns the average of the \fmacro\ scores for each class, and thus does not favour the majority class. We report the mean and standard error of \fmacro\ scores using a leave-one-out 10-fold cross-validation strategy. 

\begin{equation} \label{eq:f1macro}
    \fmacro = \frac{1}{N}\sum_{i=1}^{N}\frac{2*TP_i}{2*TP_i+FP_i+FN_i}
\end{equation}

\subsection{Large Language Model}
We conduct our experiments using LLaMa2, an open-source LLM developed by Meta AI, pretrained on a publicly available dataset of 2 trillion tokens. Specifically, we employ LLaMa2-Platypus-70B, a variant of LLaMa2 with 70 billion parameters that was fine-tuned using the Open-Platypus dataset \cite{lee2023platypus}, which focuses on enhancing the logical reasoning skills of the LLM. LLaMa2-Platypus-70B, which is fully open-source, achieved remarkable performance across several popular LLM benchmark datasets\footnote{See \url{https://huggingface.co/garage-bAInd/Platypus2-70B} for more details.}.


\subsection{Baselines}\label{sec:baselines}
We compare \textsc{Pastel} against the state-of-the-art models for the four datasets described in Table~\ref{tab:datasets}. Moreover, we include two LLM-based baselines that share the same underlying model as \textsc{Pastel} (LLaMa2), however, these baselines predict veracity directly instead of through credibility signals. Also, we distinguish between supervised and unsupervised baselines to provide a fair assessment of the methods, as supervised models are trained with access to high-quality in-domain annotated data, and thus have a significant methodological advantage over the unsupervised models. Therefore, \textbf{the supervised baselines serve as an upper bound reference for comparison against the unsupervised methods}. The baselines are described in detail below:

\subsubsection*{Unsupervised Approaches}

\begin{itemize}
    \item \textbf{Prompted weAk Supervision wiTh crEdibility signaLs (\textsc{Pastel})}: Our method, described in detail in section \ref{sec:pwscs}. The LLM extracts nineteen credibility signals for each news article in the entire dataset. We then train Snorkel's label model \cite{ratner2017snorkel} for $500$ epochs using the credibility signals extracted from the train split. Lastly, the signal weights estimated from the train set distribution are used to aggregate the signals from the test split into the final binary (\textit{misinformation}/\textit{non-misinformation}) veracity predictions.
    \item \textbf{LLaMa Zero-Shot (LLaMa-ZS)}: The LLM directly assesses the veracity of articles in the test split, without any fine-tuning. The prompt used is the same as \textsc{Pastel}'s (Figure \ref{fig:prompt}), with two slight modifications: (i) the candidate answers are `Yes' and `No', without the possibility of answering `Unsure'\footnote{Note that \textsc{Pastel} is allowed to use the `Unsure' label only for extracting the credibility signals, and not for the final veracity label.}, and (ii) the LLM answers a single question: ``Does this article contain misinformation?'', as opposed to the nineteen questions related to the credibility signals.
\end{itemize}

\subsubsection*{Supervised Approaches}

\begin{itemize}
    \item \textbf{LLaMa Fine-Tuned (LLaMa-FT)}: The LLM is fine-tuned with the causal language modeling objective using articles from the train split alongside their ground truth annotations. We employ LoRA \cite{hu2022lora} to fine-tune LLaMa2-Platypus, using the same settings as in \citet{lee2023platypus}:
    a learning rate of $3\times10^{-4}$, a batch size of $4$, and a microbatch size of $1$, and the cutoff length is set to $4096$ tokens. The training includes $100$ warmup steps, spans $1$ epoch, and employs no weight decay. The learning rate scheduler is set to cosine. For LoRA settings, we use an alpha value of $16$, a rank of $16$, and a dropout rate of $0.05$.
    Following fine-tuning, the LLM directly assesses the veracity of articles in the test split, identically to LLaMa-ZS.

    
    \item \textbf{RoBERTa}:
    As discussed in \autoref{sec:related_work}, the RoBERTa model by \citet{goel-2021-multidomain} is the state-of-the-art model for both FakeNewsAMT and Celebrity, with \fmacro\ scores of $0.99$ and $0.82$, respectively. The authors employed a single train and test split of 70\% train and 30\% test in their experimental setup, thus, we reproduce their model and evaluate it using our more robust methodology with 10-fold cross validation to ensure that the results are comparable with our other proposed baselines. We reproduce their work using the hyperparameters and settings provided in their paper: \texttt{RoBERTa-Base} pretrained model, Adam optimizer with $\beta_1$ of $0.9$ and $\beta_2$ of $0.999$, learning rate of $2e^{-5}$, weight decay of $1e^{-1}$, batch size of $8$, and $5$ training epochs.
    \item \textbf{BERT}:
    The BERT model by \citet{rai2022fake} is the state-of-the-art model for PolitiFact and GossipCop, with \fmacro\ scores of $0.88$ and $0.89$, respectively. However, we were not able to reproduce their experiments as they did not specify the hyperparameters used to finetune the model, nor did they release their code. Also, they employed a single train and test split evaluation methodology (80\% train and 20\% test), and we employ a more robust 10-fold cross validation strategy, thus, our experimental setup is not directly comparable to theirs. Therefore, we finetune a \texttt{BERT-Base-Uncased} architecture with the default hyperparameters specified in the HuggingFace deep learning framework \cite{hf-trainer}: Adam optimizer with $\beta_1$ of $0.9$ and $\beta_2$ of $0.999$, learning rate of $5e^{-5}$, batch size of $8$, and $5$ training epochs.
\end{itemize}

\vspace{0.5cm}
For experiments with LLaMa2 (LLaMa-ZS, LLaMa-FT, and \textsc{Pastel}), we apply 4-bit quantisation \cite{pmlr-v202-dettmers23a}. All experiments are conducted using a single NVIDIA A100-80GB GPU.

\section{Results} \label{sec:results}

\subsection{In-domain Classification}
In the in-domain scenario, models are trained and evaluated with in-domain data, i.e., the train and test sets are derived from the same dataset. Table~\ref{tab:main_results} presents the classification results for the proposed baselines.

\begin{table}[h]
    \centering
    \caption{Classification results (\fmacro). {\footnotesize{Highest scores for each setting are in bold. Means and standard deviations obtained with 10-fold cross-validation.}}}
    \label{tab:main_results}
    \begin{tabular}{@{}llrrrr|c@{}}
    \toprule
    \multicolumn{1}{l}{\textbf{Setting}} & \multicolumn{1}{l}{\textbf{Approach}} & \multicolumn{1}{r}{\textbf{PolitiFact}} & \multicolumn{1}{r}{\textbf{GossipCop}} & \multicolumn{1}{r}{\textbf{FNAMT}} & \multicolumn{1}{r}{\textbf{Celebrity}} & \multicolumn{1}{c}{\textbf{Mean}} \\ \midrule
    \multirow{3}{*}{Supervised} & BERT & 0.89\footnotesize{$\pm$0.03} & 0.67\footnotesize{$\pm$1.6} & 0.75\footnotesize{$\pm$0.08} & 0.79\footnotesize{$\pm$0.06} & 0.78 \\
     & RoBERTa & \textbf{0.93\footnotesize{$\pm$0.01}} & \textbf{0.80\footnotesize{$\pm$0.1}} & \textbf{0.97\footnotesize{$\pm$0.03}} & \textbf{0.87\footnotesize{$\pm$0.05}} & \textbf{0.89} \\
     & LLaMa-FT & 0.68\footnotesize{$\pm$0.02} & 0.75\footnotesize{$\pm$0.01} & 0.79\footnotesize{$\pm$0.05} & 0.43\footnotesize{$\pm$0.03} & 0.67 \\ \midrule
    \multirow{2}{*}{Unsupervised} & LLaMa-ZS & 0.61\footnotesize{$\pm$0.02} & 0.55\footnotesize{$\pm$0.01} & 0.65\footnotesize{$\pm$0.02} & 0.45\footnotesize{$\pm$0.02} & 0.57 \\
     & \textsc{Pastel} & \textbf{0.77\footnotesize{$\pm$0.01}} & \textbf{0.69\footnotesize{$\pm$0.01}} & \textbf{0.82\footnotesize{$\pm$0.01}} & \textbf{0.81\footnotesize{$\pm$0.02}} & \textbf{0.78} \\ \bottomrule
    \end{tabular}
\end{table}

First, we compare the supervised baselines: LLaMa-FT, BERT, and RoBERTa. We find that both BERT and RoBERTa significantly outperform LLaMa-FT by 0.11 and 0.22 in \fmacro, respectively, despite LLaMa-FT having a much larger number of parameters (LLaMa2 has 70 billion parameters, while BERT and RoBERTa each have fewer than 150 million parameters). This performance gap may be attributed to the relatively small size of the training data for all datasets ($<1K$ samples) except GossipCop, as larger models often require larger training sets for optimal performance \cite{NEURIPS2022_c1e2faff}. For the GossipCop dataset ($17K$ samples), LLaMa-FT outperforms BERT and is only 0.05 behind RoBERTa. Comparing the similarly sized models, BERT and RoBERTa, we find that RoBERTa, on average, outperforms BERT by 0.11 ($\uparrow 14.1\%$) in \fmacro, despite a statistical overlap (indicated by their standard deviations) in all datasets except FakeNewsAMT.

Next, we compare the unsupervised baselines, LLaMa-ZS and \textsc{Pastel}.
\textsc{Pastel} consistently outperforms LLaMa-ZS, with an average increase of 0.21 in \fmacro\ across all four datasets, which represents an average increase of 38.3\% in performance. Specifically, \textsc{Pastel} outperforms LLaMa-ZS by 0.16 ($\uparrow 22.2\%$), 0.14 ($\uparrow 25.5\%$), and 0.17 ($\uparrow 26.1\%$) for PolitiFact, GossipCop, and FakeNewsAMT, respectively. The most substantial improvement is observed for the Celebrity dataset, with an increase of 0.36 ($\uparrow 80\%$) in \fmacro. We highlight that the results obtained with the LLaMa-ZS baseline is consistent with \citet{hu2023bad} that used ChatGPT-3.5 to assess veracity for the GossipCop dataset, and obtained an \fmacro\ score of $0.57$. These results underscore \textsc{Pastel}'s substantial superiority over zero-shot prompting for veracity assessment.

Finally, we compare \textsc{Pastel} with RoBERTa, the state-of-the-art supervised model.
As discussed in Section \ref{sec:baselines}, we use the supervised models as upper bound references to \textsc{Pastel}, as they are trained with access to ground truth labels, while \textsc{Pastel} is not. Therefore, we compare RoBERTa and \textsc{Pastel} in terms of \textsc{Pastel}'s ability to approach the scores obtained by the RoBERTa model. We find that \textsc{Pastel} achieves $86.7\%$ of RoBERTa's performance averaging across the four datasets. Specifically, \textsc{Pastel} achieves $82.8\%$, $86.3\%$, $84.5\%$, and $93.1\%$ of the performance of the RoBERTa model for PolitiFact, GossipCop, FakeNewsAMT, and Celebrity, respectively.

\subsection{Cross-domain Classification}
Supervised models often experience a decline in performance when there is a mismatch between the training set distribution and the test set distribution, a phenomenon known as domain shift \cite{ben2010theory}. In this experiment, we evaluate the cross-domain robustness of the state-of-the-art supervised model, RoBERTa, in comparison to \textsc{Pastel}. For each of the four datasets $i \in D$, both models are trained with $i$, and evaluated on the three remaining datasets $j \in D\ |\ j \neq i$. Table \ref{tab:crossdataset_results} presents the cross-dataset \fmacro\ scores for both RoBERTa and \textsc{Pastel}.

\begin{table}[h]
\caption{Cross-dataset \fmacro\ for RoBERTa (\texttt{RoB}) vs. \textsc{Pastel} (\texttt{PAS}).}
\label{tab:crossdataset_results}
\begin{tabular}{@{}clcccccccc@{}}
\toprule
\multicolumn{1}{l}{} &  & \multicolumn{8}{c}{\textbf{Train}} \\
\multicolumn{1}{l}{} & \multicolumn{1}{l|}{} & \multicolumn{2}{c|}{PolitiFact} & \multicolumn{2}{c|}{GossipCop} & \multicolumn{2}{c|}{FakeNewsAMT} & \multicolumn{2}{c}{Celebrity} \\
\multicolumn{1}{l}{} & \multicolumn{1}{l|}{} & \texttt{RoB} & \multicolumn{1}{c|}{\texttt{PAS}} & \texttt{RoB} & \multicolumn{1}{c|}{\texttt{PAS}} & \texttt{RoB} & \multicolumn{1}{c|}{\texttt{PAS}} & \texttt{RoB} & \texttt{PAS} \\ \midrule
\multirow{5}{*}{\textbf{Test}} & \multicolumn{1}{l|}{PolitiFact} & x & \multicolumn{1}{c|}{x} & $0.45$ & \multicolumn{1}{c|}{$\mathbf{0.69}$} & $0.40$ & \multicolumn{1}{c|}{$\mathbf{0.67}$} & $0.65$ & $\mathbf{0.74}$ \\
 & \multicolumn{1}{l|}{GossipCop} & $0.25$ & \multicolumn{1}{c|}{$\mathbf{0.69}$} & x & \multicolumn{1}{c|}{x} & $0.21$ & \multicolumn{1}{c|}{$\mathbf{0.67}$} & $\mathbf{0.69}$ & $\mathbf{0.69}$ \\
 & \multicolumn{1}{l|}{FakeNewsAMT} & $0.54$ & \multicolumn{1}{c|}{$\mathbf{0.76}$} & $0.52$ & \multicolumn{1}{c|}{$\mathbf{0.84}$} & x & \multicolumn{1}{c|}{x} & $0.52$ & $\mathbf{0.78}$ \\
 & \multicolumn{1}{l|}{Celebrity} & $0.34$ & \multicolumn{1}{c|}{$\mathbf{0.80}$} & $0.74$ & \multicolumn{1}{c|}{$\mathbf{0.81}$} & $0.37$ & \multicolumn{1}{c|}{$\mathbf{0.78}$} & x & x \\ \cmidrule(l){2-10} 
 & \multicolumn{1}{l|}{Mean} & $0.38$ & \multicolumn{1}{c|}{$\mathbf{0.75}$} & $0.57$ & \multicolumn{1}{c|}{$\mathbf{0.78}$} & $0.33$ & \multicolumn{1}{c|}{$\mathbf{0.71}$} & $0.62$ & $\mathbf{0.74}$ \\ \bottomrule
\end{tabular}
\end{table}

On average, \textsc{Pastel} achieves a mean \fmacro\ score of $0.75$ compared to RoBERTa's $0.46$ (an increase of $63\%$). When evaluated on the PolitiFact, GossipCop, FakeNewsAMT, and Celebrity datasets, \textsc{Pastel} attains average \fmacro\ scores of $0.75$, $0.78$, $0.71$, and $0.74$, respectively. In contrast, RoBERTa achieves lower average \fmacro\ scores of $0.38$, $0.57$, $0.33$, and $0.62$ on the corresponding datasets. 

Although \textsc{Pastel} consistently outperforms RoBERTa, the difference is less pronounced for datasets within the same domain, particularly entertainment news. For instance, when RoBERTa is trained on GossipCop and tested on Celebrity, it achieves an \fmacro\ score of $0.74$, which is $0.07$ lower than \textsc{Pastel}. When trained on Celebrity and evaluated on GossipCop, both models score $0.69$. In the political domain, the performance gap is more significant. Training on PolitiFact and evaluating on FakeNewsAMT results in an \fmacro\ score of $0.54$ for RoBERTa, $0.22$ lower than \textsc{Pastel}. Similarly, training on FakeNewsAMT and testing on PolitiFact yields a score of $0.40$ for RoBERTa, which is $0.27$ below \textsc{Pastel}.

When the training and testing datasets originate from different domains, the performance difference between the models becomes more substantial. Training on political datasets and evaluating on entertainment datasets poses the most significant challenge for RoBERTa. For instance, when trained on PolitiFact and tested on GossipCop and Celebrity, RoBERTa trails \textsc{Pastel} by $0.44$ and $0.46$, respectively. Similar gaps are observed when training on FakeNewsAMT and testing on these datasets, with RoBERTa falling behind by $0.46$ on GossipCop and $0.41$ on Celebrity. This trend persists when training on entertainment news and testing on political news, albeit with a smaller gap between the two models. Training on GossipCop and testing on PolitiFact and FakeNewsAMT results in gaps of $0.24$ and $0.32$, respectively. Training on Celebrity and testing on the same two datasets results in gaps of $0.09$ and $0.26$.

These results underscore the superior robustness of \textsc{Pastel} to domain shift compared to the supervised state-of-the-art model. This characteristic is crucial for applications where in-domain training data is unavailable, or for dynamically changing domains and emergent topics. 

\subsection{Error Analysis}
\begin{figure}[h]
    \centering
    \includegraphics[width=0.22\textwidth]{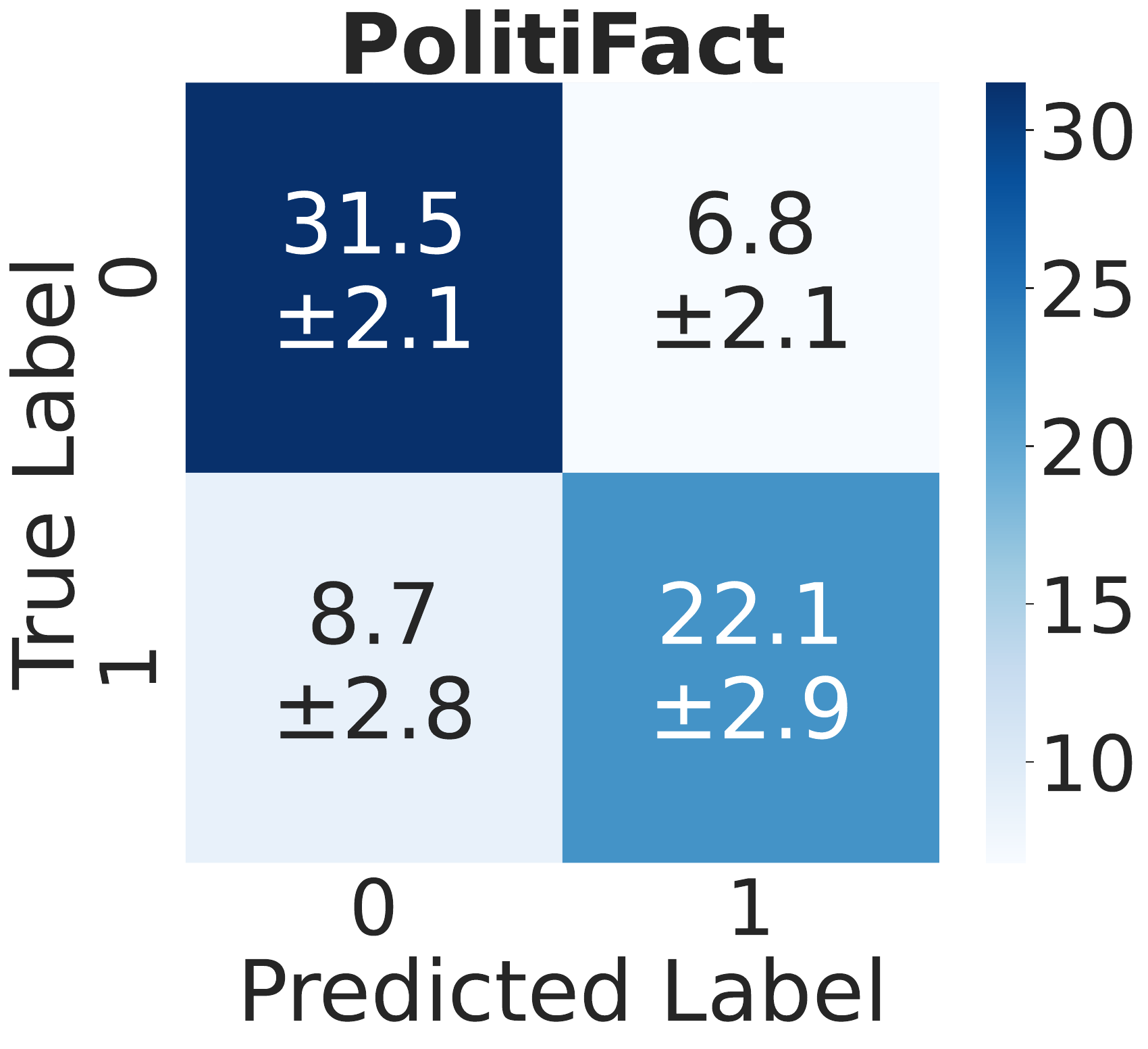}
    \hfill
    \includegraphics[width=0.24\textwidth]{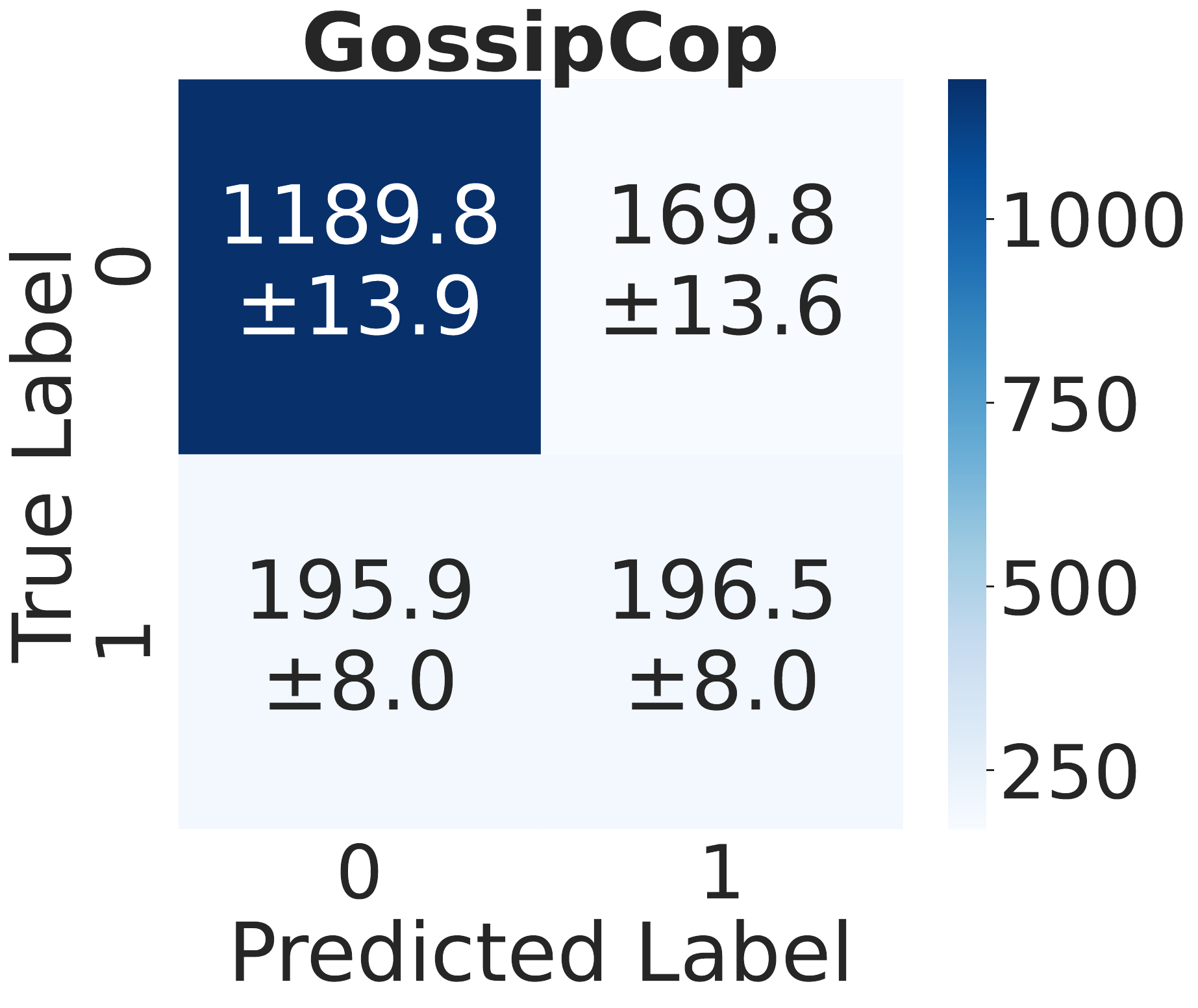}
    \hfill
    \includegraphics[width=0.22\textwidth]{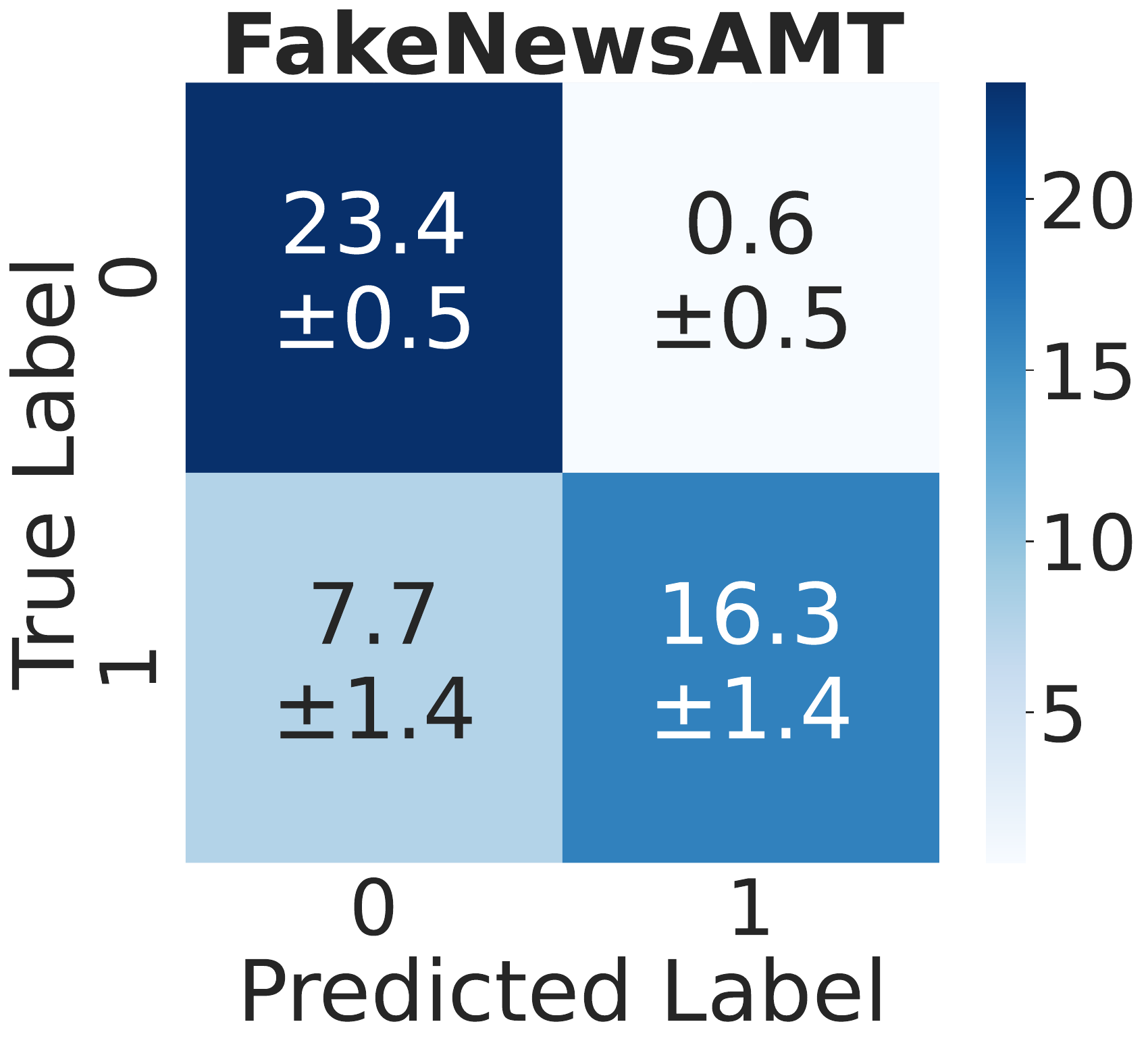}
    \hfill
    \includegraphics[width=0.22\textwidth]{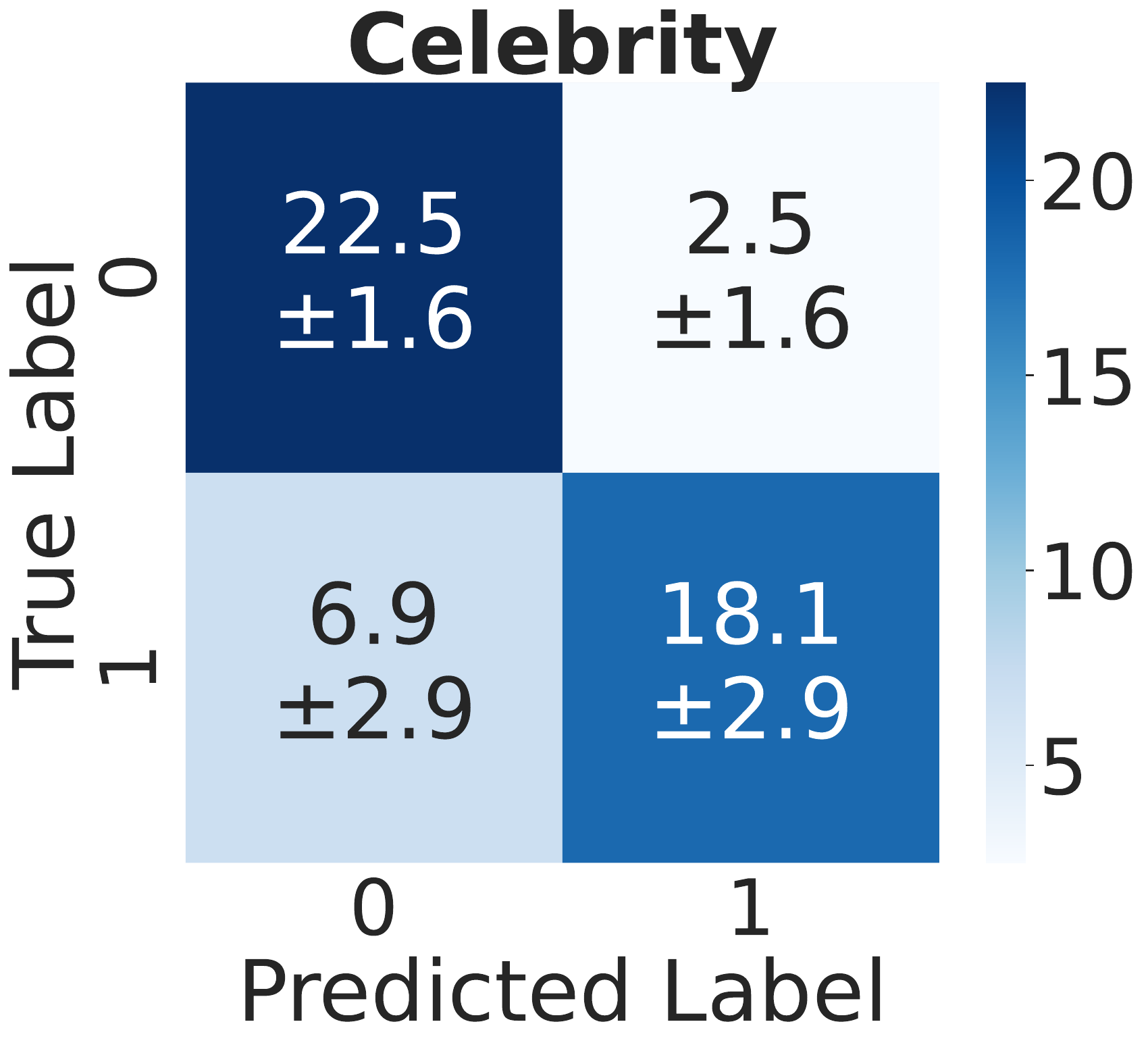}
    \caption{Mean confusion matrices obtained with \textsc{Pastel}. Means and standard deviations reported across 10-fold cross-validation. Labels $0$ and $1$ refer to \textit{non-misinformation} and \textit{misinformation}, respectivelly.}
    \label{fig:confusion_matrices}
\end{figure}

To gain deeper insights into the performance of our method, we conduct a detailed error analysis to systematically identify the types of errors made by \textsc{Pastel}. \autoref{fig:confusion_matrices} displays the confusion matrices averaged over 10-fold cross-validation for test sets in each dataset.

As each dataset has different sizes and label distributions, we further calculate the False Positive Rate (FPR) and the False Negative Rate (FNR) (see Equation~\ref{eq:fnr_fpr}). Table~\ref{tab:fnr_fpr} displays the FNR and FPR for each dataset.


\begin{equation}
\text{FNR} = \frac{\text{FN}}{\text{TP} + \text{FN}} \qquad \text{FPR} = \frac{\text{FP}}{\text{FP} + \text{TN}}
\label{eq:fnr_fpr}
\end{equation}

\vspace{0.5cm}


\begin{table}[h]
    \centering
    \caption{False Negative Rate (FNR) and False Positive Rate (FPR).}
    \begin{tabular}{lcc}
        \toprule
        \textbf{Dataset} &\textbf{ FNR (\%)} & \textbf{FPR (\%)} \\
        \midrule
        PolitiFact & 28.2 & 17.7 \\
        GossipCop & 49.3 & 13.2 \\
        Celebrity & 21.6 & 10.0 \\
        FakeNewsAMT & 32.1 & 2.4 \\ \midrule
        Mean & 32.8 & 10.8 \\
        \bottomrule
    \end{tabular}
    \label{tab:fnr_fpr}
\end{table}

 Across all datasets, \textsc{Pastel} yields a higher rate of false negatives over false positives, with averages of $32.8\%$ and $10.8\%$, respectively.
 \textsc{Pastel}'s FNR is notably high for the GossipCop dataset ($49.3\%$), which is possibly a result of its label skewnewss, as the negative class comprises $77.6\%$ of the dataset. Contrastingly, the FNR for the other three datasets is considerably lower, with $32.1\%$, $28.2\%$, and $21.5\%$ for FakeNewsAMT, PolitiFact, and Celebrity, respectively.

In the context of \textsc{Pastel}'s method, false negative errors occur when one or more signals are not triggered. To examine such errors, we compare the distribution of credibility signals in true positive (TP) and false negative (FN) examples. In Table~\ref{tab:error_analysis_signals}, we present the relative frequency (the number of times the credibility signal was triggered, divided by the number of articles) of each credibility signal in TP and FN predictions, averaged across the four datasets.

\begin{table*}[!h]
\centering
\caption{Relative frequency of credibility signals triggered in True Positive (TP) and  False Negative (FN) predictions. Percent decrease indicated within parenthesis.}
\label{tab:error_analysis_signals}
\begin{tabular}{r|ll}
\hline
\textbf{Credibility Signal} & \textbf{TPs} & \textbf{FNs} \\ \hline
Emotional Valence            & 0.52 & 0.00\textsubscript{(↓100\%)} \\
Clickbait                    & 0.29 & 0.00\textsubscript{(↓100\%)} \\
Expert Citation              & 0.55 & 0.00\textsubscript{(↓100\%)} \\
Evidence                     & 0.56 & 0.00\textsubscript{(↓100\%)} \\
Source Credibility           & 0.25 & 0.00\textsubscript{(↓100\%)} \\
Bias                         & 0.39 & 0.01\textsubscript{(↓97.4\%)} \\
Document Citation            & 0.73 & 0.02\textsubscript{(↓97.3\%)} \\
Incivility                   & 0.26 & 0.01\textsubscript{(↓96.2\%)} \\
Sensationalism               & 0.69 & 0.03\textsubscript{(↓95.7\%)} \\
Polarising Language          & 0.39 & 0.02\textsubscript{(↓94.9\%)} \\
Misleading about content     & 0.57 & 0.05\textsubscript{(↓91.2\%)} \\
Explicitly Unverified Claims & 0.32 & 0.03\textsubscript{(↓90.6\%)} \\
Incorrect Spelling           & 0.19 & 0.02\textsubscript{(↓89.5\%)} \\
Impoliteness                 & 0.08 & 0.01\textsubscript{(↓87.5\%)} \\
Informal Tone                & 0.34 & 0.08\textsubscript{(↓76.5\%)} \\
Personal Perspective         & 0.38 & 0.09\textsubscript{(↓76.3\%)} \\
Reported by Other Sources    & 0.78 & 0.38\textsubscript{(↓51.3\%)} \\
Call to Action               & 0.04 & 0.02\textsubscript{(↓50.0\%)} \\
Inference                    & 0.24 & 0.14\textsubscript{(↓41.7\%)} \\ \hline
Total                        & 7.57 & 0.91\textsubscript{(↓88.0\%)} \\ \hline
\end{tabular}
\end{table*}

The statistics indicate that all $19$ signals occur less frequently in FN predictions compared to TP. On average, $7.57$ credibility signals are triggered in TP predictions, whereas only $0.91$ signals are triggered in FN predictions, representing a significant decrease of $88.0\%$. A reduction of more than $70\%$ in frequency is observed for $16$ signals, while \textit{Reported by Other Sources}, \textit{Call to Action}, and \textit{Inference} show smaller decreases of $51.3\%$, $50.0\%$, and $41.7\%$, respectively.

\section{Analysis of Credibility Signals} \label{sec:signals_analyses}
This section examines the effectiveness of LLM-extracted credibility signals in predicting content veracity through two research questions: (i) Is there a statistical association between credibility signals and the article's veracity?
(ii) Which credibility signals contribute the most towards \textsc{Pastel}'s classification performance? 


\subsection{Credibility Signals and Veracity} \label{sec:distribution}

\begin{figure}[!h]
    \centering
    \begin{adjustbox}{width=\linewidth}
        \includegraphics{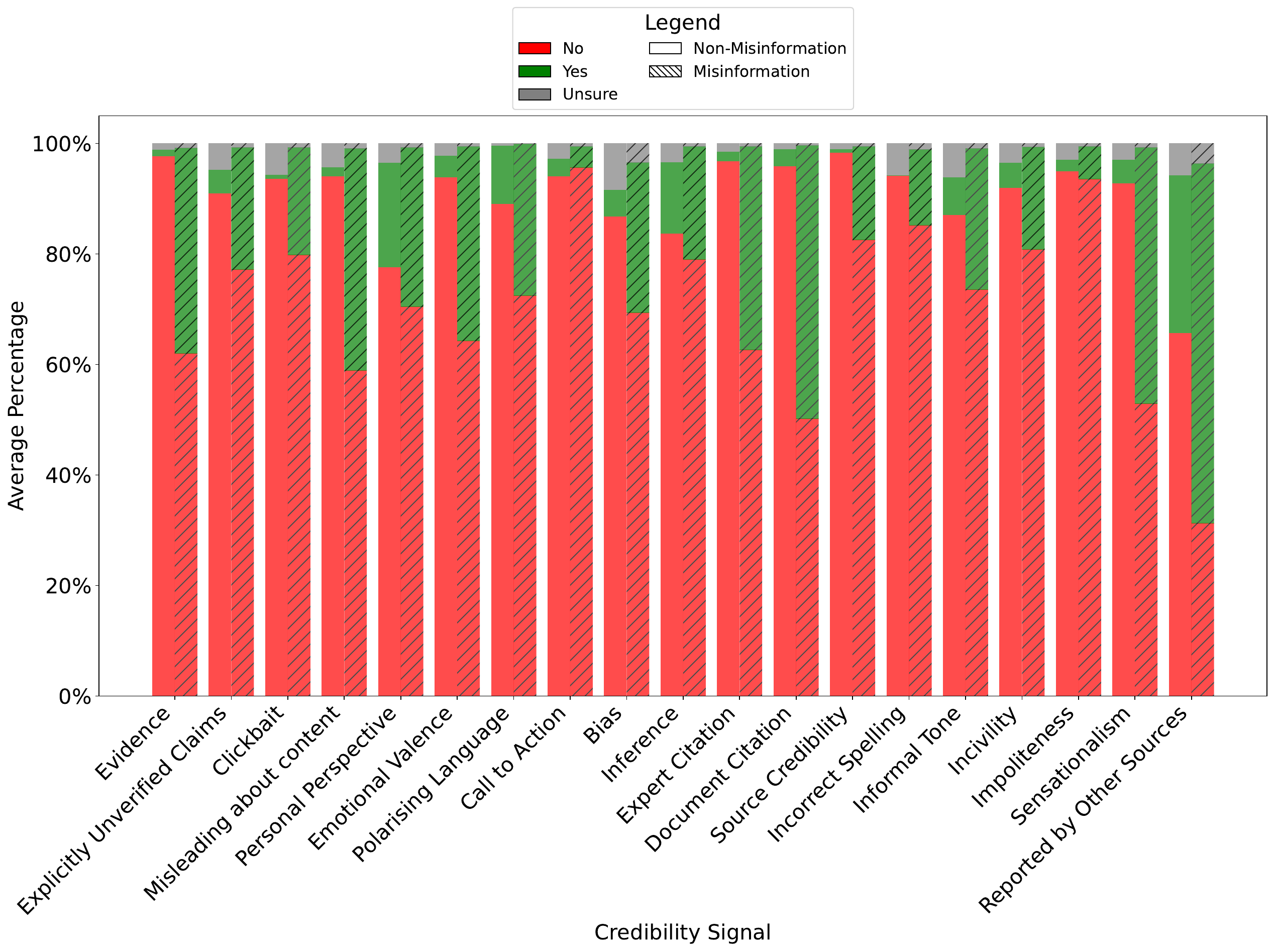}
        \end{adjustbox}
        \caption{Distribution of LLM responses per credibility signal for \textit{non-misinformation} articles (solid bars) and \textit{misinformation} articles (hashed bars) averaged across all datasets.}
        \label{fig:avg_signal_distribution}
    \end{figure}

Figure~\ref{fig:avg_signal_distribution} compares the proportion of LLM responses (`Yes', `No', or `Unsure') for each credibility signal in \textit{misinformation} and \textit{non-misinformation} articles.

Firstly, we note that the percentage of `Unsure' answers is relatively small across all credibility signals, composing less than $10\%$ of the answers. Also, the rate of `Unsure' answers is higher for \textit{non-misinformation} articles. These statistics may indicate that the model is overconfident, or in other words, is often not capable of identifying when there is not enough information to confidently decide between `Yes' or `No'. Nevertheless, all $19$ credibility signals are found more frequently in \textit{misinformation} articles than in \textit{non-misinformation} articles.

\begin{figure}[h]
\centering
\begin{adjustbox}{width=\linewidth}
  \includegraphics{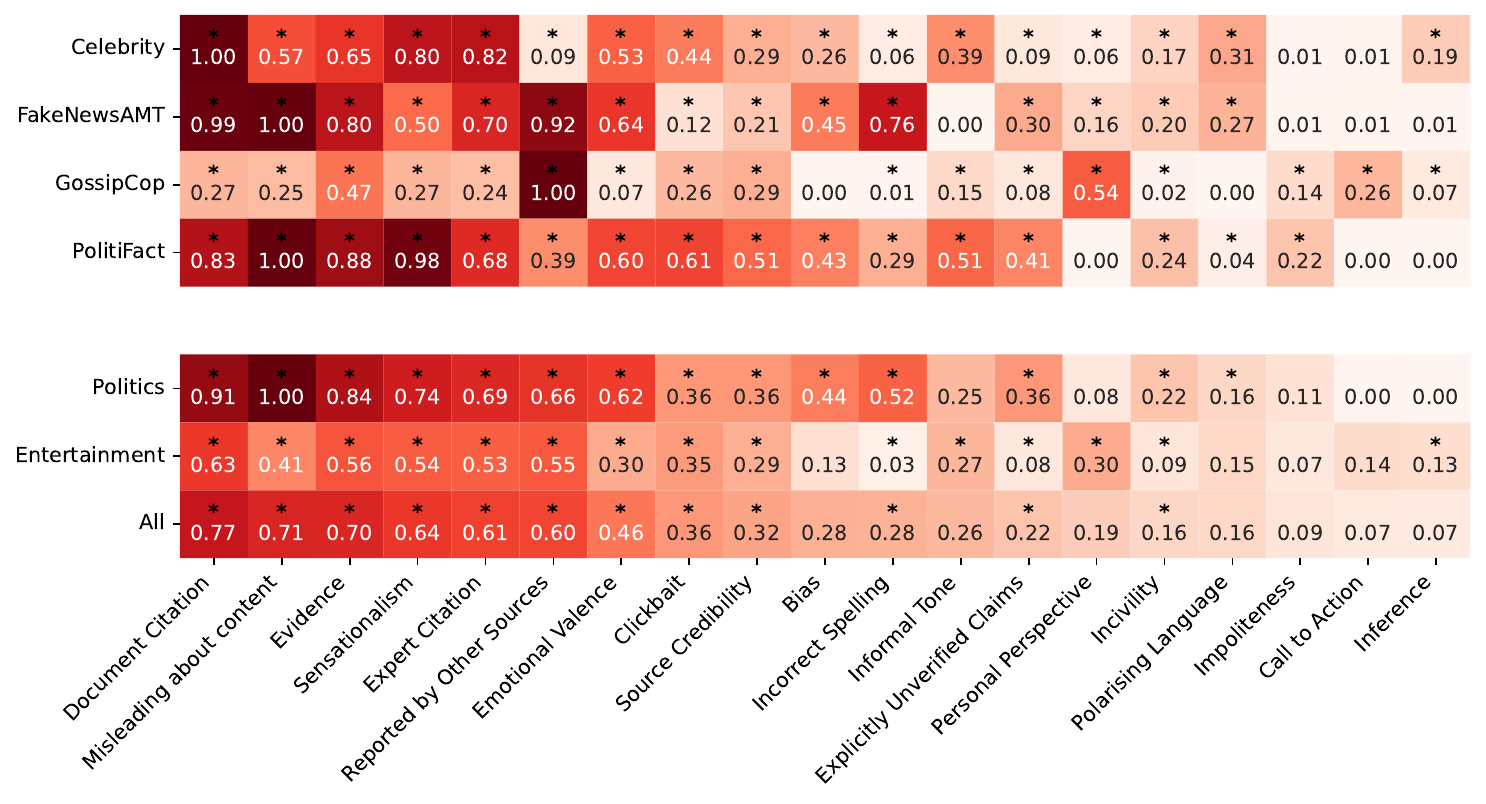}
\end{adjustbox}
  \caption{Normalised Pearson's $\chi^2$ statistics per credibility signal. Credibility signals where the null hypothesis $H_0$ is rejected ($p<0.05$, $1$ degree of freedom) are marked with an asterisk ($\ast$). Results are shown for each dataset, and aggregated by domain; `Politics' displays the average of FakeNewsAMT and PolitiFact, and `Entertainment' shows the average of Celebrity and GossipCop. All four datasets are averaged into `All'. For aggregate results, we reject $H_0$ if $H_0$ is rejected in all the aggregated datasets. Credibility signals are sorted in descending order based on the overall average (`All').}
  \label{fig:chi2}
\end{figure}

In order to verify if there is a statistically significant association between the credibility signals and the article's veracity, we perform a Pearson's chi-squared statistical test. Our null hypothesis $H_0$ is that there is no association between the credibility signals and the veracity of the article. We reject the null hypothesis $H_0$ if $p<0.05$. This test is done for each credibility signal independently, and for each dataset separately. Additionally, we analyse the $\chi^2$ statistic as a measure of the strength of association between the credibility signal and the veracity label. A higher $\chi^2$ statistic suggests a significant deviation in the observed distribution of a given signal between \textit{misinformation} and \textit{non-misinformation} articles. For ease of visualisation, the $\chi^2$ statistics are normalised between $0$ and $1$. Figure~\ref{fig:chi2} illustrates the test outcomes.


When averaging across all datasets (All), we reject $H_0$ for $12$ credibility signals, that therefore have a statistically significant association with the veracity of the articles across all four datasets: \textit{Document Citation}, \textit{Misleading about content}, \textit{Evidence}, \textit{Sensationalism}, \textit{Expert Citation}, \textit{Reported by Other Sources}, \textit{Emotional Valence}, \textit{Clickbait}, \textit{Source Credibility}, \textit{Incorrect Spelling}, \textit{Explicitly Unverified Claims}, and \textit{Incivility}. Out of these $12$ signals, $6$ display a particularly high average normalised $\chi^2$ ($\geq0.6$), indicating a strong association: \textit{Document Citation}, \textit{Misleading about content}, \textit{Evidence}, \textit{Sensationalism}, \textit{Expert Citation}, and \textit{Reported by Other Sources}. For the other remaining $6$ signals, $H_0$ is rejected only within specific domains. For instance, $H_0$ is rejected for the signals of \textit{Inference}, \textit{Personal Perspective}, and \textit{Informal Tone} in the Entertainment domain, but not in Politics. Conversely, we only reject $H_0$ for the signals \textit{Polarising Language} and \textit{Bias} in the Politics domain. Lastly, for some signals, $H_0$ is rejected only in specific datasets: \textit{Impoliteness} for GossipCop and PolitiFact, and \textit{Call to Action} for GossipCop. 

In conclusion, all $19$ signals show a statistically significant association with the article's veracity in at least one dataset, with the majority ($12$ signals) demonstrating a strong association across all four datasets. Additionally, domain-specific signals exist where $H_0$ is only rejected within either the Politics or Entertainment domains, but not both.


\subsection{Ablation Study} \label{sec:ablation}
In this experiment, we evaluate the contribution of each credibility signal to \textsc{Pastel}’s performance through an ablation study. We iteratively remove each of the $19$ credibility signals from the dataset, training the label model on the remaining $18$ signals. We then compare the performance of this modified model against the model trained with all $19$ signals. Table~\ref{tab:ablation} shows the percentage change in \fmacro\ when each signal is excluded.

\begin{table*}[!ht]
\caption{Ablation study results. Scores are the percentage change in performance when a certain credibility signal is excluded from the dataset. Signals are sorted increasingly by the mean score.}
\centering
\tiny
\label{tab:ablation}
\begin{tabular}{@{}l|rrrr|rr|r@{}}
\toprule
\textbf{Signal Removed} & \textbf{PolitiFact} & \textbf{GossipCop} & \textbf{FNAMT} & \textbf{Celebrity} & \textbf{Entert.} & \textbf{Politics} & \textbf{Mean} \\ \midrule
Document Citation            & -0.6 & -1.0 & -1.4 & -1.5 & -1.2 & -1.0 & -1.1 \\
Sensationalism               & -0.6 & -0.5 & -0.5 & -2.0 & -1.2 & -0.5 & -0.9 \\
Misleading about content     & -0.4 & -0.3 & -2.7 & 0.6  & 0.2  & -1.6 & -0.7 \\
Incorrect Spelling           & -0.2 & 0.1  & -1.7 & 0.0  & 0.0  & -0.9 & -0.4 \\
Clickbait                    & -0.3 & -0.1 & 0.0  & -0.4 & -0.3 & -0.2 & -0.2 \\
Informal Tone                & -0.6 & 0.0  & 0.2  & -0.5 & -0.2 & -0.2 & -0.2 \\
Source Credibility           & -0.2 & -0.1 & 0.0  & 0.0  & 0.0  & -0.1 & -0.1 \\
Explicitly Unverified Claims & -0.6 & 0.2  & -0.5 & 0.7  & 0.4  & -0.6 & -0.1 \\
Impoliteness                 & 0.0  & 0.0  & 0.0  & -0.3 & -0.2 & 0.0  & -0.1 \\
Expert Citation              & -0.5 & -0.2 & 0.6  & 0.1  & -0.1 & 0.0  & 0.0  \\
Call to Action               & 0.4  & 0.0  & 0.0  & -0.3 & -0.1 & 0.2  & 0.0  \\
Inference                    & 0.5  & -0.1 & 0.0  & -0.2 & -0.2 & 0.3  & 0.1  \\
Reported by Other Sources    & 0.5  & 0.0  & 0.0  & 0.3  & 0.2  & 0.3  & 0.2  \\
Incivility                   & 0.9  & 0.2  & -0.3 & 0.0  & 0.1  & 0.3  & 0.2  \\
Bias                         & 0.9  & 0.1  & 0.0  & 0.0  & 0.1  & 0.5  & 0.3  \\
Personal Perspective         & 1.6  & 0.2  & -0.3 & 0.0  & 0.1  & 0.7  & 0.4  \\
Emotional Valence            & 1.6  & 0.2  & 0.1  & -0.2 & 0.0  & 0.8  & 0.4  \\
Evidence                     & -0.3 & -0.3 & 0.8  & 1.3  & 0.5  & 0.2  & 0.4  \\
Polarising Language          & 2.0  & 0.3  & -0.3 & 0.3  & 0.3  & 0.8  & 0.6  \\ \bottomrule
\end{tabular}
\end{table*}

Overall, individual signals exhibit a relatively small impact on the model's performance. The most influential signal, \textit{Document Citation}, reduces the model’s performance by an average of $1.1\%$ across all datasets. The top nine signals positively impacting performance, i.e., those that lead to lower \fmacro\ scores when removed, are: \textit{Document Citation}, \textit{Sensationalism}, \textit{Misleading about content}, \textit{Incorrect Spelling}, \textit{Clickbait}, \textit{Informal Tone}, \textit{Source Credibility}, \textit{Explicitly Unverified Claims}, and \textit{Impoliteness}. Except for \textit{Informal Tone} and \textit{Impoliteness}, all show a statistically significant association with veracity (see Figure~\ref{fig:chi2}).

In contrast, eight signals reduce \textsc{Pastel}’s performance on average, as indicated by an increase in \fmacro\ scores when removed. These, in descending order of impact, are: \textit{Polarising Language}, \textit{Evidence}, \textit{Emotional Valence}, \textit{Personal Perspective}, \textit{Bias}, \textit{Incivility}, \textit{Reported by Other Sources}, and \textit{Inference}. Despite their negative average effect, some signals demonstrate domain-specific benefits, such as \textit{Expert Citation}, \textit{Call to Action}, and \textit{Inference} in Entertainment, and \textit{Misleading about content}, \textit{Incorrect Spelling}, \textit{Source Credibility}, and \textit{Explicitly Unverified Claims} in Politics.

These findings underscore that \textsc{Pastel}’s strength lies in its ability to aggregate multiple credibility signals, as no single signal significantly affects the overall performance on its own. Although some signals, such as \textit{Document Citation} and \textit{Sensationalism}, demonstrate utility across multiple domains, the degree of effectiveness of credibility signals is often domain-specific. For example, while signals like \textit{Source Credibility} and \textit{Misleading about Content} improve performance primarily in the political domain, others such as \textit{Expert Citation} and \textit{Call to Action} show benefits in entertainment.

\section{Discussion and Conclusion} \label{sec:discussion_conclusion}
In this work, we proposed \textsc{Pastel}, a novel approach that uses LLMs to extract a wide range of credibility signals, which are then aggregated with weak supervision to predict veracity. Extensive experiments show that \textsc{Pastel} significantly outperforms the unsupervised baseline (LLaMa-ZS) by $38.3\%$. Additionally, \textsc{Pastel} achieves $86.7\%$ of the performance of the supervised state-of-the-art RoBERTa model by \citet{goel-2021-multidomain}, without using any form of human supervision (neither labelled data nor user interactions as in previous work \cite{shu2020weaksocialsupervision, helmstetter2018weakly, Wang_Yang_Ma_Xu_Zhong_Deng_Gao_2020}). In cross-domain classification, \textsc{Pastel} outperforms the supervised state-of-the-art model by a large margin ($63\%$). These results demonstrate the usefulness of our method mainly in  scenarios where no in-domain labelled data is available.

\textsc{Pastel}’s ability to leverage credibility signals in a zero-shot setting enables it to maintain high performance across diverse domains, making it well-suited for dynamically changing environments and emergent topics. For example, during the early stages of the COVID-19 pandemic in late 2019, misinformation about the virus spread rapidly, while labelled datasets for training supervised models were not available until mid to late 2020 \cite{info:doi/10.2196/19273,hossain-etal-2020-covidlies, cui2020coaid}. Additionally, \textsc{Pastel} offers a key advantage over other weakly supervised methods for misinformation detection that rely on user interactions \cite{shu2020weaksocialsupervision, helmstetter2018weakly, Wang_Yang_Ma_Xu_Zhong_Deng_Gao_2020}. These approaches depend on users engaging with harmful content before detection is possible, by which time the misinformation may have already caused significant damage.  In contrast, \textsc{Pastel} operates directly at the content level, allowing it to detect misinformation in its early stages of dissemination.

We studied the association between the LLM-predicted credibility signals and the human-annotated veracity labels, revealing that $12$ out of the $19$ signals exhibit a statistically significant association across all four datasets. Moreover, we observed domain-specific credibility signals that demonstrate higher degrees of association with datasets related to Politics compared to Entertainment, and vice versa. This finding can guide future work in crafting more specialised sets of credibility signals for specific domains. Next, we conducted an ablation study to measure the contribution of each credibility signal towards \textsc{Pastel}’s performance in predicting veracity. We found that the contribution of individual signals is relatively small, and that \textsc{Pastel}’s performance depends on the collective influence of it's wide range of credibility signals rather than in one signal in specific.

Plenty of research opportunities arise from the implications of this work. Future research may explore the usefulness of multi-modal credibility signals. For instance, the report by \citet{w3c-signals} describes credibility signals associated with images, such as the originality of the photo and whether it has been manipulated or not. Signals related to audio, video, and even the structure of the content, such as the ads presented, can be considered. Another promising research direction is to explore and mitigate the overconfidence of the LLM when extracting credibility signals, as seen in Figure~\ref{fig:avg_signal_distribution}, where the LLM seldom responds with `Unsure’, which can degrade performance.

\section*{Ethical considerations}
LLMs are known to inherit biases from their training data \cite{nadeem-etal-2021-stereoset}, which can manifest in their interpretations and judgements regarding the presence or absence of credibility signals in textual content. These biases may lead to inaccuracies or disparities in signal detection, potentially favouring certain types of content or perspectives over others. Moreover, the deployment of LLM-based systems in real-world applications must navigate concerns around fairness, transparency, and accountability. Researchers and developers are therefore urged to mitigate biases through rigorous testing, data preprocessing, and continuous monitoring.

Also, although efforts aimed at mitigating misinformation are crucial in combating its harmful effects, it is important to acknowledge that these efforts can inadvertently empower malicious actors \cite{10.1145/3581783.3612704}. By gaining insights into which credibility signals are more easily detected by LLMs, and which correlate more strongly with veracity, malicious users could potentially exploit this knowledge to enhance their misinformation tactics and circumvent automatic detection systems. Therefore, we strongly urge researchers to apply our methodology with caution and in accordance with best practice ethics protocols.

\subsection*{Abbreviations}
PASTEL, Prompted Weak Supervision with Credibility Signals; LLM, Large Language Model; NLP, Natural Language Processing; QA, Question-Answering; CWCG, Credible Web Community Group; W3C, World Wide Web Consortium; PWS, Programmatic Weak Supervision; TP, True positives; TN, True negatives; FP, False positives; FN, False negatives; FNR, False negative rate; FPR, False positive rate; ZS, Zero-shot; FT, Fine-tuned; RoBERTa, Robustly Optimized BERT Approach; BERT, Bidirectional Encoder Representations from Transformers; PAS, PASTEL; RoB, RoBERTa, FNAMT, FakeNewsAMT; Entert., Entertainment.

\section*{Declarations}
\subsection*{Availability of data and materials}
Our code to reproduce the experiments is made fully available at \url{https://github.com/JAugusto97/PASTEL}. The datasets used in the experiments are publicly available: (1) FakeNewsAMT and Celebrity \cite{perez-rosas-etal-2018-automatic} (\url{https://lit.eecs.umich.edu/downloads.html}), and (2) PolitiFact and GossipCop \cite{shu2020fakenewsnet} (\url{https://github.com/KaiDMML/FakeNewsNet}).

\subsection*{Competing interests}
The authors declare that they have no competing interests.

\subsection*{Funding}
This  work  has  been  co-funded  by  the  UK’s innovation agency (Innovate UK) grant 10039055 (approved under the Horizon Europe Programme as vera.ai,  EU  grant  agreement  101070093) under action number 2020-EU-IA-0282. João Leite is supported by a University of Sheffield EPSRC Doctoral Training Partnership (DTP) Scholarship.

\subsection*{Authors' contributions}
JL developed the PASTEL method, participated in the literature review, performed the experiments, and wrote the manuscript. OR participated in the literature review. OR, CS and KB participated in writing the manuscript and provided direction with conceptualisation and methodology. All authors edited and submitted the final manuscript. All authors read and approved the final manuscript.

\subsection*{Acknowledgements}
We thank Freddy Heppel, Ivan Srba, and Ben Wu for their valuable feedback. We also acknowledge IT Services at The University of Sheffield for the provision of services for High Performance Computing.

\clearpage

\bibliography{sn_article}

\end{document}